\documentclass{bmvc2k}

\title{Edge-aware Bidirectional Diffusion for\\Dense Depth Estimation from Light Fields}
\addauthor{Numair Khan}{numair_khan@brown.edu}{1}
\addauthor{Min H. Kim}{minhkim@kaist.ac.kr}{2}
\addauthor{James Tompkin}{james_tompkin@brown.edu}{1}

\addinstitution{
 Brown University
}
\addinstitution{
 KAIST
}

\runninghead{Khan et al.}{Edge-aware Bi-directional Diffusion}

\usepackage{amsmath,amssymb,graphicx}
\usepackage{bbm}
\usepackage{multirow}
\usepackage{booktabs}
\usepackage{array}
\usepackage{hhline}
\usepackage{paralist}
\usepackage[mathscr]{eucal}
\usepackage[inline]{enumitem}

\usepackage[subtle]{savetrees}

\begin{document}

\maketitle

\begin{abstract}
We present an algorithm to estimate fast and accurate depth maps from light fields via a sparse set of depth edges and gradients. Our proposed approach is based around the idea that true depth edges are more sensitive than texture edges to local constraints, and so they can be reliably disambiguated through a bidirectional diffusion process. First, we use epipolar-plane images to estimate sub-pixel disparity at a sparse set of pixels. To find sparse points efficiently, we propose an entropy-based refinement approach to a line estimate from a limited set of oriented filter banks. Next, to estimate the diffusion direction away from sparse points, we optimize constraints at these points via our bidirectional diffusion method. This resolves the ambiguity of which surface the edge belongs to and reliably separates depth from texture edges, allowing us to diffuse the sparse set in a depth-edge and occlusion-aware manner to obtain accurate dense depth maps. 
\emph{Project webpage: \href{http://visual.cs.brown.edu/lightfielddepth}{http://visual.cs.brown.edu/lightfielddepth}}
\end{abstract}

\section{Introduction}
\label{sec:introduction}

\begin{figure}[t]
\begin{minipage}[c]{1.0\linewidth}
    \centering
    \includegraphics[width=0.3\linewidth,clip,trim={0cm 11.4cm 6cm 0.3cm}]{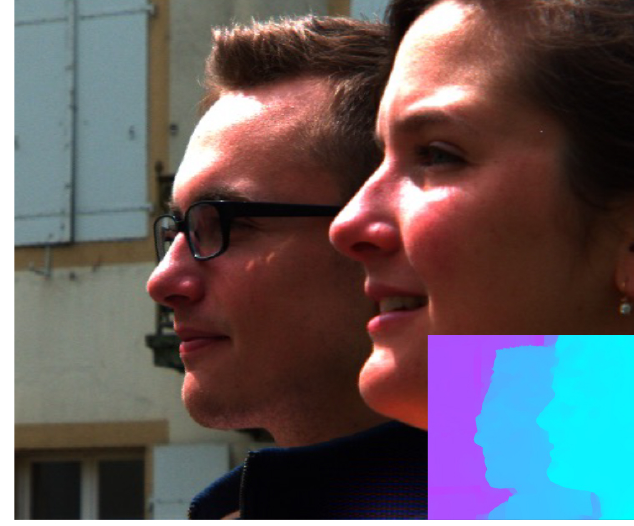}
    \includegraphics[width=0.3\linewidth,clip,trim={0cm  5.7cm 6cm 6cm}]{./Figures/editing.pdf}
    \includegraphics[width=0.3\linewidth,clip,trim={0cm    0.1cm 6cm 11.6cm}]{./Figures/editing.pdf}
\end{minipage} \\
\begin{minipage}[c]{1.0\linewidth}
    \caption[]{Light field editing requires high-accuracy depth edges. \textit{Middle:} The dense depth estimated by Zhang et al.~\cite{zhang2016} illustrates the difficulty of extracting correct edges, with inserted content appearing at incorrect depths or with overhanging regions. \textit{Right}: The accuracy of our depth edges allows effective occlusion handling when editing light fields. The inset shows our disparity map.} 
    \label{fig:editing1}
\end{minipage}
\vspace{-0.35cm}
\end{figure}

Light fields record small view changes onto a scene. This allows them to store samples from both the spatial and angular distributions of light. The additional angular dimension allows imaging applications such as synthetic aperture photography and view interpolation~\cite{wilburn2005,isaksen2000}. Most of these applications can be directly implemented in image space using image-based rendering (IBR) techniques~\cite{levoy1996,gortler1996}. For applications such as light field editing and augmented reality, we require an explicit scene representation in the form of a point cloud, depth map, or derived 3D mesh, to allow occlusion-aware and view-consistent processing, editing, and rendering.

However, light field depth estimation is a difficult problem. 
Oftentimes, state-of-the-art methods strive for geometric accuracy without always considering occlusion edges, which are especially important for handling visibility in light field editing applications. 
Further, while the many views allow dense and accurate depth to be derived, the extra angular dimension carries large data costs that makes most depth estimation algorithms computationally inefficient~\cite{jeon2015,zhang2016}. Recent methods have sought to overcome this barrier by learning data-driven priors with deep learning. 
While this can be effective, it requires additional training data, and may overfit to scenes or capture scenarios~\cite{li2020}. 




We present a first-principles method for estimating occlusion-accurate depth maps from light fields with no learned priors and demonstrate its application in light field editing tasks. This is achieved by estimating disparity at a sparse set of pixels identified as most important for the final result. These estimates are then propagated to all pixels using occlusion-aware diffusion. Traditionally, diffusion pipelines for depth completion attempt to recover a complete description of depth maps via a sparse set of depth edges and gradients~\cite{elder1999}. Commonly, techniques follow three steps~\cite{holynski2018,yucer2016}:
\begin{enumerate}[topsep=1ex,itemsep=0.1ex,partopsep=1.5ex,parsep=1ex]
\item Obtain sparse depth labels accurately and efficiently,
\item Determine diffusion gradient at each labeled point, and
\item Perform dense depth diffusion.
\end{enumerate}

Step 1 is critical yet difficult: finding sparse depth labels via edges in EPIs requires robustness to noise and occlussion-awareness. For this, we identify unwanted edges by observing gradients along and just next to the edge. Second, using large filter banks for subpixel depth precision is expensive. Thus, from an initial depth estimate from a moderately-sized filter bank, we propose a novel entropy-based depth refinement using efficient random search 
to obtain a subpixel estimate.

Step 2 is \emph{also} critical yet difficult: determining the diffusion direction requires us to know the depth at pixels around each label, but for efficiency we only have a sparse set of labeled points. Holynski and Kopf~\cite{holynski2018} deal with this by assuming that sparse labels do not lie on depth edges so that neighboring pixels have a similar label. Yucer et al.~\cite{yucer2016} handle labels on depth edges, but their method is designed for light fields with a large number ($\approx$3000+) of views. Our novel contribution here is that we determine diffusion direction from other sparse labels within context via a bidirectional `backward-forward' diffusion process. Together, improvements in these steps allow fast and accurate occlusion estimation for light fields. 





\vspace{-0.2cm}
\section{Related Work}
\label{sec:relatedwork}
\vspace{-0.1cm}
The information implicit within an EPI (Epipolar-Plane Image) is useful for depth or disparity estimation algorithms, and the regular structure of an EPI obviates the need for extensive angular regularization. Thus, many light field operations seek to exploit it~\cite{park2017robust}. Khan et al.~\cite{khan2019} use a set of large Prewitt filters to reliably detect oriented lines in EPIs, then diffuses these across all light field views using occlusion-aware edges to guide a depth inpainting process~\cite{khan2020vclfd}. However, their estimate of which edges are depth edges can be inaccurate, leading to errors in diffusion. Zhang et al.'s~\cite{zhang2016} spinning parallelogram operator works in a related fashion on EPIs, but has a larger support than a Prewitt filter. It provides accurate depth estimates, but is slow to execute. Their approach is similar to To\v{s}i\'{c} and Berkner's~\cite{tosic2014} convolution with a set of specially adapted kernels to create light field scale-depth spaces. Wang et al.~\cite{wang15, wang16} build on this by proposing a photo-consistency measure to address occlusion. Tao et al.'s \cite{tao2013} work considers higher dimensional representations of EPIs which allows them to use both correspondence and defocus to get depth. 

The relation between defocus and depth is also exploited by the sub-pixel cost volume of Jeon et al.~\cite{jeon2015}, who also present a method for dealing with the distortion induced by micro-lens arrays. An efficient and accurate method for wide-baseline light fields was proposed by Chuchwara et al.~\cite{chuchvara2020}. They use an oversegmentation of each view to get initial depth proposals, which are iteratively improved using PatchMatch~\cite{barnes09}. 
Also related to our method is the work of Holynski and Kopf~\cite{holynski2018}, who present an efficient method for depth densification from a sparse set of points for augmented reality applications. However, they assume that the set of sparse points and their depth values are known beforehand. Our method does not make this assumption, and seeks to identify both the points and their depth as well as performing dense diffusion. Yucer et al.~\cite{yucer2016} present a diffusion-based method that uses image gradients to estimate a sparse label set. However, their method is designed to work for light fields with thousands of views. Chen et al.~\cite{chen2018} estimate accurate occlusion boundaries from superpixels to regularize the depth estimation process. In general, densification methods~\cite{xu2019, wang2018, cheng2018} largely seek to recover accurate metric depth without considering occlusion boundaries.

Many methods have sought to use data-driven methods to learn priors to avoid the cost of dealing with a large number of images, and to overcome the loss of spatial information induced by the spatio-angular tradeoff in lenslet images. Huang et al.'s~\cite{huang18} work can handle an arbitrary number of uncalibrated views. Alperovich et al.~\cite{alperovich2018} use an encoder-decoder architecture to perform an intrinsic decomposition of a light field, and also recover disparity for the central cross-hair of views. Jiang et al.~\cite{jiang2018,jiang2019} fuse the disparity estimates at four corner views estimated using a deep learning optical-flow method, and Shi et al.~\cite{shi2019} build on this by adding a refinement network to the fusion pipeline. Li et al.~use oriented relation networks to learn depth from local EPI analysis~\cite{li2020}. In general, learned prior methods have been successful in estimating depth~\cite{cheng2018, shin18, xu2019, eldesokey2019}; we show that a method without any learned priors or training data requirements can be efficient and effective.

\begin{figure}[t]
    \centering
    \includegraphics[width=0.45\linewidth,clip,trim={0.5cm 11.5cm 0.5cm 0cm}]{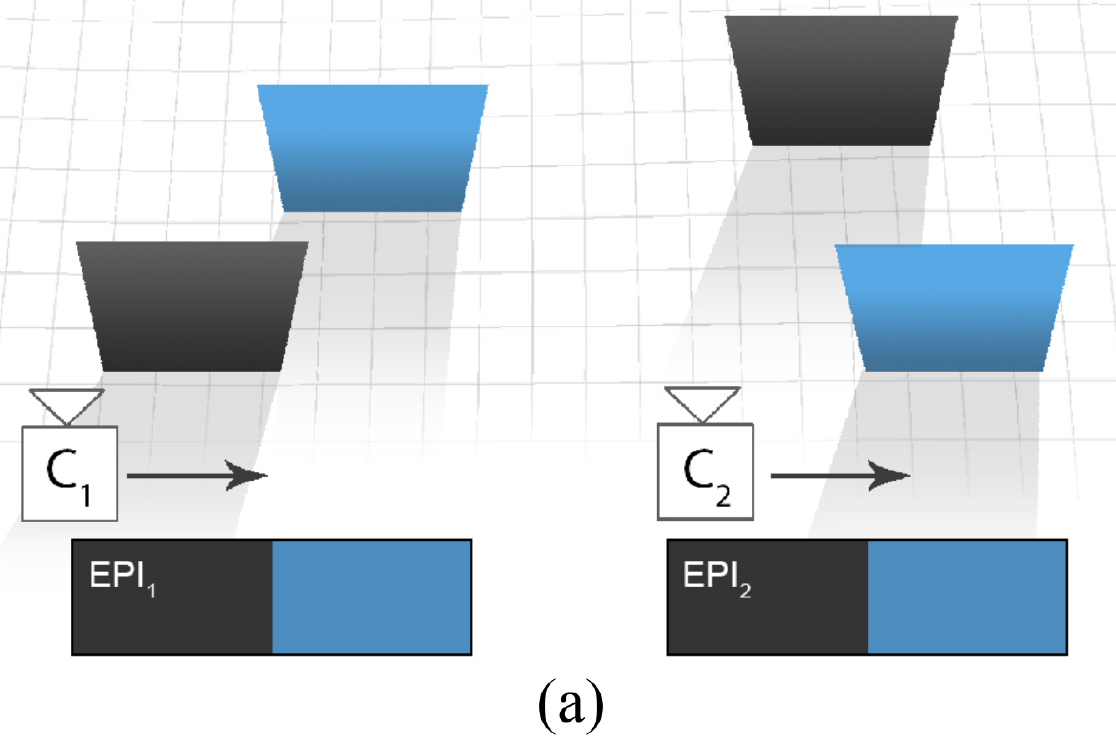}
    \hspace{4mm}
    \includegraphics[width=0.35\linewidth,clip,trim={0cm 0.1cm 0cm 8.3cm}]{Figures/non-local-edges.pdf}
    \vspace{-3mm}
    \caption{Sparse labels at edges are difficult to propagate because the edge is weakly localized at the boundary of two projected surfaces. As a result, labels may be assigned to the incorrect side of a depth boundary. \textbf{(a)} Two different scene configurations captured with cameras $C_1$ and $C_2$ may generate similar EPIs. The EPI edge represents the boundary of the occluding surface. For $C_1$ this is the surface on the left (black);  for $C_2$ it is on the right (blue). \textbf{(b)} The direction from which occlusion happens cannot be disambiguated from edge activations alone, leading to incorrect label placement.}
    \label{fig:epi-depth-ambiguity}
    \vspace{-4mm}
\end{figure}
\vspace{-0.2cm}
\section{Our Approach}
\label{sec:method}
\vspace{-0.1cm}

Our goal is to estimate disparity at a sparse set of points such that their labels can be efficiently diffused to generate occlusion-accurate depth maps. Based on this requirement we populate our sparse set for diffusion by selecting points around light field edges (Section~\ref{sec:method-sparse-labels}). However, while past work on image reconstruction has shown that edges are sufficient for recovering a perceptually accurate representation of the original image~\cite{elder1999}, labels at edges are poorly localized at the intersection of surfaces (Fig.~\ref{fig:epi-depth-ambiguity}). Hence, we use a bi-directional diffusion process to determine the propagation direction that generates the most accurate occlusion boundaries (Section~\ref{sec:method-diffusion-gradient-estimation}).

\subsection{Sparse Depth Labels from EPI Edges}
\label{sec:method-sparse-labels}

\begin{figure*}[t]
\centering
\includegraphics[width=\linewidth]{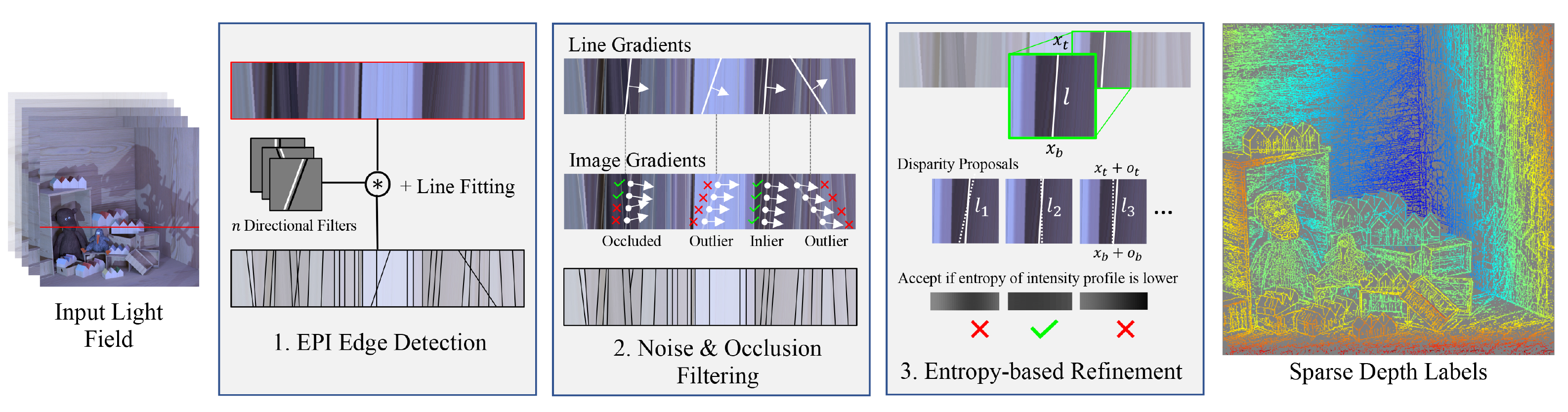}%
\vspace{-3mm}
\caption{EPI edges provide both the location and disparity labels of a sparse point set $\mathcal{P}$. Thus, the first stage of our sparse labeling pipeline consists of EPI edge detection and line fitting. In the second stage, we compare the direction of each EPI line with underlying image gradients to remove noisy labels and points that are occluded in the central view. Finally, we improve the disparity estimates of the sparse set through an entropy-based random search. }
\label{fig:sparse-labeling}
\end{figure*}

An EPI (Epipolar-Plane Image) provides an angular slice through a 4D light field, and has a linear structure resulting from epipolar geometry constraints: points in world space become lines in an EPI, with the slope of each line corresponding to the depth of the point. The regularity of an EPI makes it easy to identify salient edges and their depth at the same time. 

\paragraph{Noise \& Occlusion Filtering.}
For each EPI $I$ in the central cross-hair of views, we use large Prewitt filters~\cite{khan2019} to recover a set $\mathcal{L}$ of parametric lines representing edge points in 4D space. This process, while fast, tends to generate many false-positives. To filter these out we use a gradient-based alignment scheme: each line $l \in \mathcal{L}$ is sampled at $n$ locations to generate the set of samples $S_l = \{(x_i, y_i)\}$. The line $l$ is considered a false-positive if the local image gradient of $I$ does not align with the line direction at a minimum $k$ number of samples:
\begin{align}
        \sum_{\mathrm{\textbf{s}} \in S_l} \mathbbm{1}\left( \frac{\nabla I(\mathrm{\textbf{s}}) (\nabla l)^T}{\lVert \nabla I(\mathrm{\textbf{s}}) \rVert \lVert \nabla l \rVert} > cos(\tau_f) \right) < k,
\label{eqn:epi-outlier-rejection}
\end{align}
where $\mathbbm{1}(\cdot)$ is the indicator function that counts the set of aligned samples, $\nabla I$ is the first-order image gradient approximated using a 3$\times$3 Sobel filter, and $\nabla l$ is perpendicular to the line. The parameters $\tau_f$ and $k$ are constants with $\tau_f = \pi/13$ and $k=(\mathrm{EPI\ height})/c$, with $1\leq c\leq \mathrm{EPI\ height}$. To determine the constant value $c$, we consider two factors: 
\begin{inparaenum}[1)]
\item the accuracy of EPI line fitting, and 
\item the expected minimum number of views a point is visible in.
\end{inparaenum}
In the case of perfect alignment between the line and EPI gradients, $c=1$. This means that a line with even a single misaligned sample is rejected. However, if a point is occluded in some views, the corresponding EPI line will be hidden and misalignment of samples in those views is inevitable. If we set $c = 1$ we risk discarding such lines. We determine empirically that $c=4$ provides good results across the synthetic and real world scenes, and across the narrow and wider baseline light fields that we evaluate on.

The parametric definition of EPI lines does not carry any visibility information for a point across light field views. We determine visibility $v(l)$ of a point $l \in \mathcal{L}$ in the central view as:
\begin{align}
    v(l) = \mathbbm{1}\left( \frac{ \nabla  I (\mathrm{\textbf{s}}_c) (\nabla l)^T}{\lVert \nabla I(\mathrm{\textbf{s}_c}) \rVert \lVert \nabla l \rVert} > cos(\tau_v)\right),
\end{align}
where $\mathrm{\textbf{s}}_c$ is the EPI sample corresponding to the central view and $\tau_v = \pi/10$. 

\paragraph{Entropy-based Disparity Refinement.}
Notice that the number of discrete disparity values of points in $\mathcal{L}$ is bounded by the number of large Prewitt filters used for EPI line fitting. Computational efficiency considerations prevent this number from becoming too large. Moreover, numerical precision and sampling errors result in the granularity of depth estimates plateauing beyond a certain number of filters. Thus, to enable the calculation of sub-pixel disparity values we fine-tune the initial estimates through random search and filtering. Let  $\mathcal{L}_c = \{l \in \mathcal{L}~|~v(l) = 1\}$. Then 
for each $l \in \mathcal{L}_c$ and image samples $S_l = \{(x_i, y_i)\}$ along the line we minimize the energy function defined by the entropy of normalized intensity values: 
\begin{align}
    E(l) = \sum_{\mathrm{\textbf{s}} \in S_l} -P(I(\mathrm{\textbf{s}}))\log_2(P(I(\mathrm{\textbf{s}}))),
\end{align}
where $I(\mathrm{\textbf{s}})$ is the intensity value at $\mathrm{\textbf{s}}$ and $P(\mathrm{\textbf{s}})$ is estimated from a histogram. 

We minimize $E(l)$ by performing a random search in the 2D parameter space defined by the x-intercepts of $l$ on the top and bottom edge of the EPI, $l = (x_t, x_b)$: at the $j$th iteration of the search we generate uniform random numbers $(o_t, o_b) \sim U(-1, 1)(\alpha t^j)$, to generate a proposal $l_{j} = (x_t + o_t, x_b + o_b)$ (Fig.~\ref{fig:sparse-labeling}). This is accepted with probability one if $E(l_j) < E(l_{j-1})$. We use $t =$ 0.88, $\alpha =$ 0.15 and run the search for a maximum of 10 iterations. 

The resulting disparity estimates are then refined by joint filtering in the spatial, disparity, and LAB color space. Let $\mathcal{P}$ represent the spatial projection of $\mathcal{L}_c$ into the central view, and let $p_s$, $p_d$, and $p_c$ be the spatial position, disparity, and color of a point $p \in \mathcal{P}$. The filtered disparity estimate $f(p_d)$ is calculated via a spatial neighborhood $\mathscr{S}$ around $p$:
\begin{align}
   f(p_d) = \frac{1}{W} \sum_{q \in \mathscr{S}} \mathcal{N}_{\sigma_s}(\lVert p_s - q_s \rVert) \mathcal{N}_{\sigma_d}(p_d - q_d) \mathcal{N}_{\sigma_c}(\lVert p_c - q_c \rVert)p_d,  \nonumber
\end{align}
where the normalization factor $W$ is given by
\begin{align}
   W = \sum_{q \in \mathscr{S}} \mathcal{N}_{\sigma_s}(\lVert p_s-q_s \rVert) \mathcal{N}_{\sigma_d}(p_d - q_d) \mathcal{N}_{\sigma_c}(\lVert p_c - q_c \rVert).
\end{align}

We found that the combination $\sigma_s=$~10, $\sigma_d=$~0.1 and $\sigma_c=$~0.5 works for all scenes.

\vspace{-0.2cm}
\subsection{Occlusion Edges via Diffusion Gradients}
\label{sec:method-diffusion-gradient-estimation}
\vspace{-0.1cm}

\begin{figure}[t]
\centering
\subfigure[]{\raisebox{2.5mm}{\includegraphics[width=0.17\textwidth]{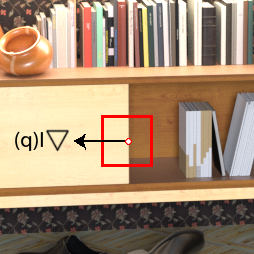}}}
\subfigure[]{\includegraphics[width=0.39\textwidth]{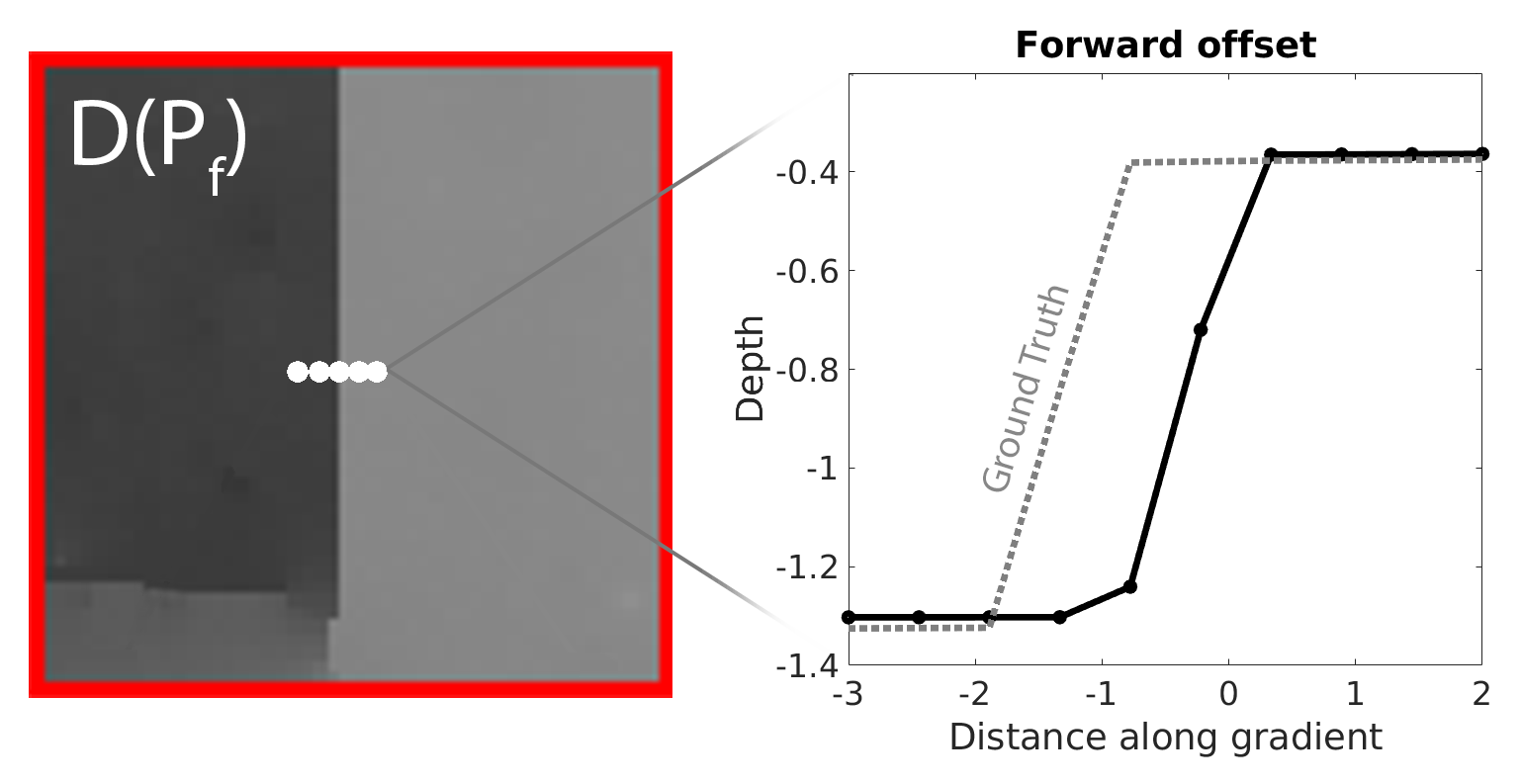}}
\subfigure[]{\includegraphics[width=0.39\textwidth]{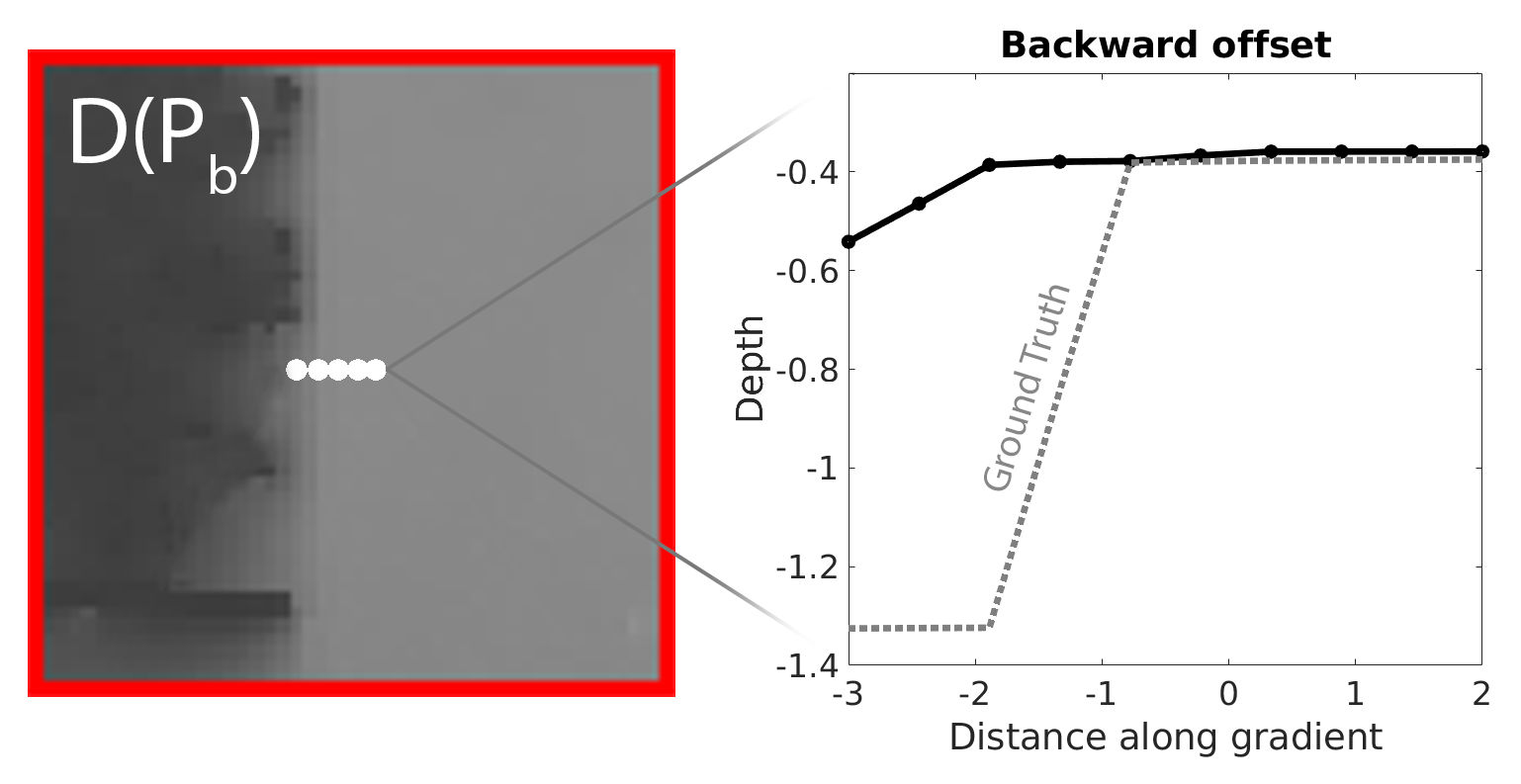}}
\caption{\textbf{(a)} Given an edge point $p$ with image gradient $\nabla I(p)$ and depth label $p_d$ we would like to determine which side of the edge to propagate $p_d$. We generate images $\hat D[\mathcal{P}_f]$ (\textbf{b}), and $\hat D[\mathcal{P}_b]$ (\textbf{c}) by solving a Poisson optimization problem with diffusion direction $p + \nabla I(p)$ and $p - \nabla I (p)$ respectively. The correct diffusion direction (\textbf{b}) generates an intensity profile resembling a step function. In the example shown, $p_d$ corresponds to the surface on the right of the edge as $p + \nabla I(p)$ generates a profile more closely resembling a step function.}
\label{fig:offset-direction-curves}
\vspace{-0.25cm}
\end{figure}



We want to diffuse the sparse set of disparity labels in $\mathcal{P}$ to a dense grid of pixels $\hat{D}$ such that $\nabla \hat{D}$ accurately represents all occlusion edges in the scene. However, this is a chicken-and-egg problem as we need the occlusion edges to determine the diffusion direction at each $p~\in~\mathcal{P}$. As Figure \ref{fig:epi-depth-ambiguity} shows, the disparity for a point lying on an EPI edge alone is not sufficient to determine the surface direction in which to perform diffusion. Directly propagating the sparse disparity estimates to generate a dense depth map results in significant errors around edges (supplemental Fig.~\ref{fig:epi-naive-diffusion}). 

As all potential occlusion edges are also depth edges, one way to determine diffusion direction is by distinguishing depth and texture edges. Yucer et al.~\cite{yucer2016} do this by comparing the variation in texture on both sides of an edge as the view changes: the background seen around a depth edge will change more rapidly than the foreground, leading to a larger variation in texture along one side of the edge. The correct diffusion direction is to the side with lower variation. This method works for light fields with thousands of views (3000+ images), but proves ineffective on datasets that are captured using a lenslet array or camera rig (Fig.~\ref{fig:results-edges}). 
This is because the assumption fails to hold in cases where 
\begin{enumerate*}
    \item the background lacks texture, and
    \item the light field has a small baseline with relatively few views, which is common for handheld cameras. Here, occlusion is minimal and image intensity variation is caused more by sensor noise than by background texture variation.
\end{enumerate*}

Our proposed solution to the depth edge identification problem works for light fields with few views (e.g., 7$\times$7 from a Lytro). We use $S[\mathcal{A}]$ to represent the image created by splatting sparse points in a set $\mathcal{A}$ onto a $w\times h$ raster grid, and $D$ to be a dense $w\times h$ disparity map. Diffusion is formulated as a constrained quadratic optimization problem:
\begin{align}
  \hat D[\mathcal{A}] = \underset{D}{\mathrm{argmin}} \sum_{p\in \mathcal{A}} E_d(p) + \sum_{(p, q)\in \mathscr{S}} E_{s}(p, q),
  \label{eqn:diffusion-general}
\end{align}
where $\hat D[\mathcal{A}]$ is the optimal disparity map given the sparsely labeled image $S[\mathcal{A}]$ and $\mathscr{S}$ is the set of all four-connected neighbors in $D$. The data term $E_d(p)$ and smoothness term $E_s(p, q)$ are defined as:
\begin{align}
  E_d(p) = \lambda_d(p) \big\lVert S[\mathcal{A}](p) - D(p) \big\rVert,  
  \hspace{0.5cm}\text{and}\hspace{0.5cm}
  E_s(p, q) = \lambda_s(p) \big\lVert D(p) - D(q) \big\rVert,
  \label{eqn:prop-constraint-data} 
\end{align}
with $\lambda_d(\cdot)$ and $\lambda_s(\cdot)$ being the spatially-varying data and smoothness weights.

Equation~\eqref{eqn:diffusion-general} represents a standard Poisson problem, and we solve it using an implementation of the LAHBPCG solver~\cite{szeliski2006} by posing the constraints in the gradient domain as proposed by Bhat et al.~\cite{bhat2009}. We begin by defining two sets formed from opposite offset directions $\nabla I(p)$ and $-\nabla I(p)$:
\begin{align}
    \mathcal{P}_f = \{ p + \nabla I(p)~\forall~p \in \mathcal{P}\},
    \hspace{0.5cm}\text{and}\hspace{0.5cm}
    \mathcal{P}_b = \{ p - \nabla I(p)~\forall~p \in \mathcal{P}\},
\end{align}%
where $\nabla I(p)$ is the gradient of the central light field view at point $p$. Then, we solve Equation~\eqref{eqn:diffusion-general} for both offset directions $\hat D[\mathcal{P}_f]$ and $\hat D[\mathcal{P}_b]$ using data and smoothness weights:
\begin{align}
    \lambda_d(p) = \left\{ \begin{array}{ll}
    10^6 & \mbox{if $p \in \mathcal{A}$,} \\
    0 & \mbox{otherwise,} \end{array} \right.
    \hspace{0.5cm}\text{and}\hspace{0.5cm}
    \lambda_s(p) = \frac{1}{\lVert \nabla I(p) \rVert + \epsilon}.
\end{align}

\begin{figure}[t]
\centering
\includegraphics[width=0.8\linewidth]{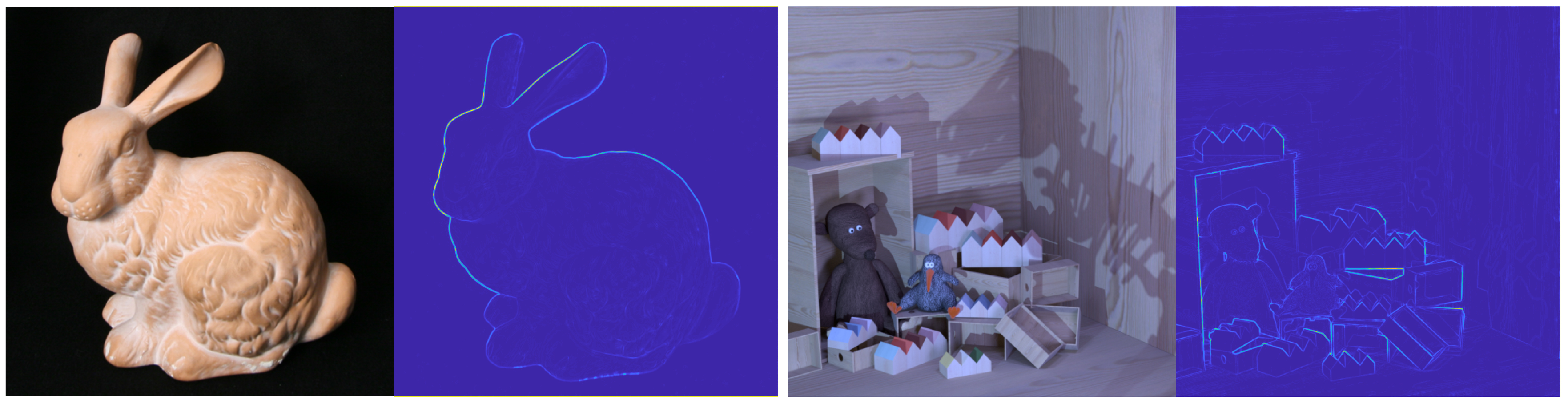}
\vspace{-0.25cm}
\caption{\emph{Estimated depth edge confidence $\lambda_s$.} The bi-directional diffusion process allows us to identify depth edges by considering the mean gradient at each pixel across the backward-forward pass. Texture edge gradient remains low in both passes. For depth edges, the gradient is higher in one pass. For depth edges that are not meant to be sharp, the change in depth around that region from the bi-directional solve is small, and picking either offset leads to low error.
}
\label{fig:epi-depth-edges}
\vspace{-0.40cm}
\end{figure}

Given both solutions, we compare the normalized depth profile around each point $p \in P$ along $\nabla I(p)$ in $\hat D[\mathcal{P}_f]$ and $\hat D[\mathcal{P}_b]$. Figure~\ref{fig:offset-direction-curves} shows that the profile for the correct offset direction ($\nabla I(p)$ or $-\nabla I(p)$) more closely resembles a step function around $p$ due to a strong depth gradient. This is because neighboring points in the correct offset direction will have a disparity value similar to $p$. The high data term together with the global smoothness constraint results in a small gradient around $p$ when the incorrect offset pushes it to the wrong side of the edge. We estimate the profile around $p$ in $\hat D[\mathcal{P}_f]$ and $\hat D[\mathcal{P}_b]$ by convolving the normalized value of a set $N_p$ of pixels around $p$ with the step filter $\mathrm{\textbf{F}} = [{-1}~{-1}~1~1]$. We define:
\begin{align}
  \lambda_e(p) =  \underset{\{\hat D[\mathcal{P}_f], \hat D[\mathcal{P}_b]\}}{\mathrm{max}} \lVert N_p \circledast \mathrm{\textbf{F}} \rVert.
\end{align}
The final map $\hat D[ \mathcal{Q}]$ with the desired depth edges is generated using Equation~\eqref{eqn:diffusion-general} where $\mathcal{Q} = \{ p \pm \nabla I(p) ~\forall~p \in \mathcal{P}\}$ is a sparse set of points offset in the diffusion direction determined above. The final data and smoothness weights are:
\vspace{-0.3cm}
\begin{eqnarray}
    \label{eqn:diffusion-data-final}
    \lambda_d(p) = \omega~ \mathrm{exp}(a\lambda_e(p)),
    \hspace{0.5cm}\text{and}\hspace{0.5cm}
    \lambda_s(p) = \frac{1}{\lVert \nabla I(p) \rVert \lVert \nabla \hat D[ \mathcal{P}_f] + \nabla \hat D[\mathcal{P}_b] \rVert},
\end{eqnarray}
where $\lambda_s(p)$ defines the depth edge confidence at every pixel (Fig.~\ref{fig:epi-depth-edges}). The parameters in Equation~\eqref{eqn:diffusion-data-final} are set as $\omega=1.5\times 10^2$ and $a=3$. These values work for all scenes.

\vspace{-0.1cm}
\section{Experiments}
\label{sec:experiments}
\vspace{-0.1cm}

\begin{figure*}[p]
\centering
\includegraphics[width=0.85\textwidth]{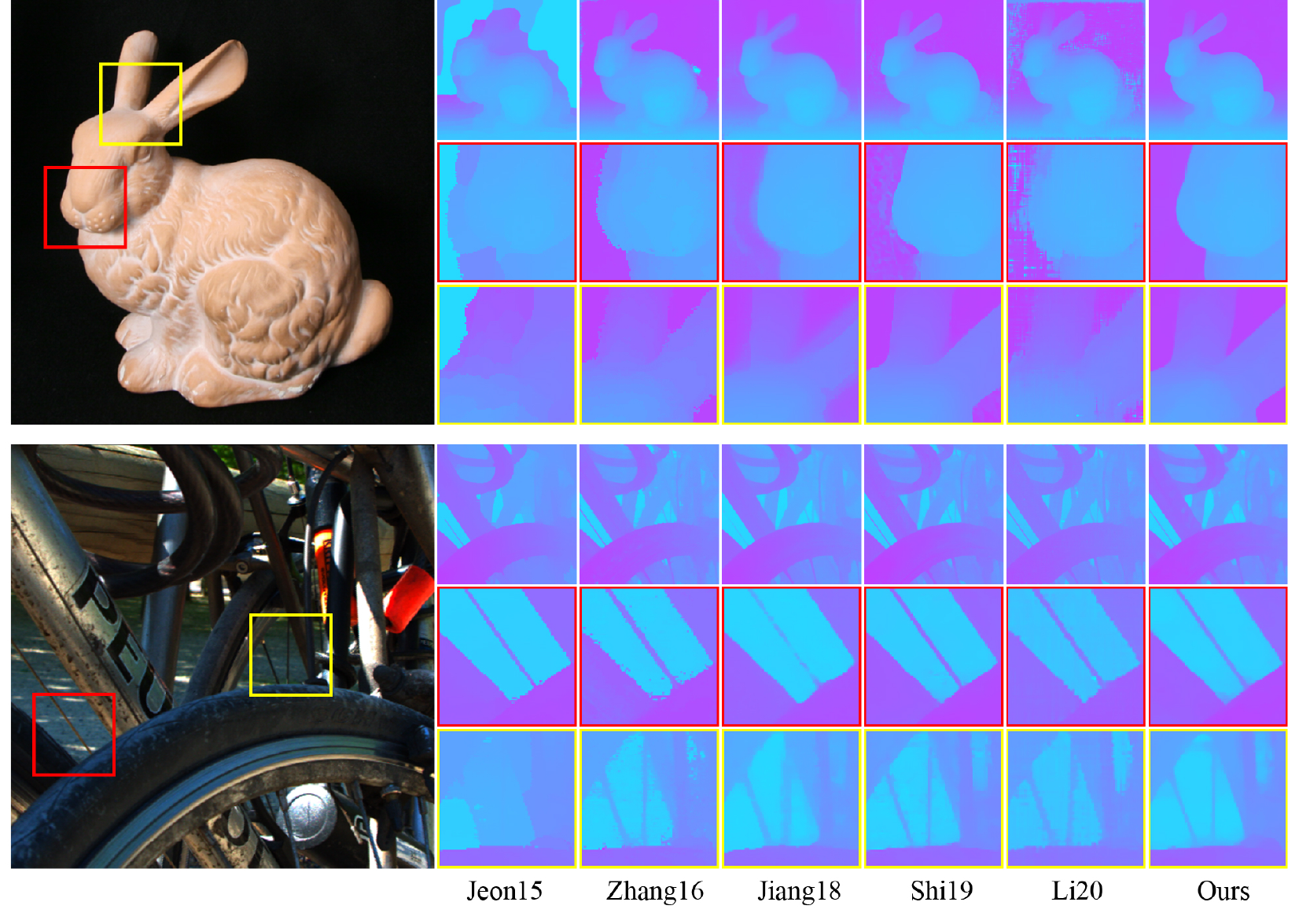}%
\vspace{-4mm}
\caption{Occlusion edges in disparity maps. \emph{Top:} Stanford dataset light field captured with a camera rig. \emph{Bottom:} EPFL light field from a Lytro Illum. \emph{Left to right:} Jeon et al.~\cite{jeon2015}, Zhang et al~\cite{zhang2016}, Jiang et al.~\cite{jiang2018}, Shi et al.~\cite{shi2019}, and ours. } 
\label{fig:results-real}
\vspace{-3mm}
\end{figure*}

\paragraph{Occlusion Edge Accuracy.}
Qualitatively, our method produces sharper and more accurate occlusion edges than state-of-the-art light field depth estimation methods. We compare our results to three non-learning-based methods: the defocus and correspondence cues methods by Jeon et al.~\cite{jeon2015} and Wang et al.~\cite{wang16}, and the spinning parallelogram operator of Zhang et al.~\cite{zhang2016}. We also compare with the learning-based methods of Jiang et al.~\cite{jiang2018}, Shi et al.~\cite{shi2019}, and Li et al.~\cite{li2020}. We do not compare to Holynski and Kopf~\cite{holynski2018}: this uses COLMAP, which fails on typical skew-projected light field data.
In Figure~\ref{fig:results-real}, we show results on light fields from the EPFL MMSPG Light-Field Dataset~\cite{rerabek2016} (7$\times$7) and the Stanford Light Field Archive~\cite{stanford2008} (17$\times$17). The latter dataset is captured with a camera rig and has a wider baseline than the EPFL light fields, which come from a Lytro Illum camera. 

\begin{figure*}[p]
\vspace{-3mm}
\centering
\includegraphics[width=0.85\textwidth]{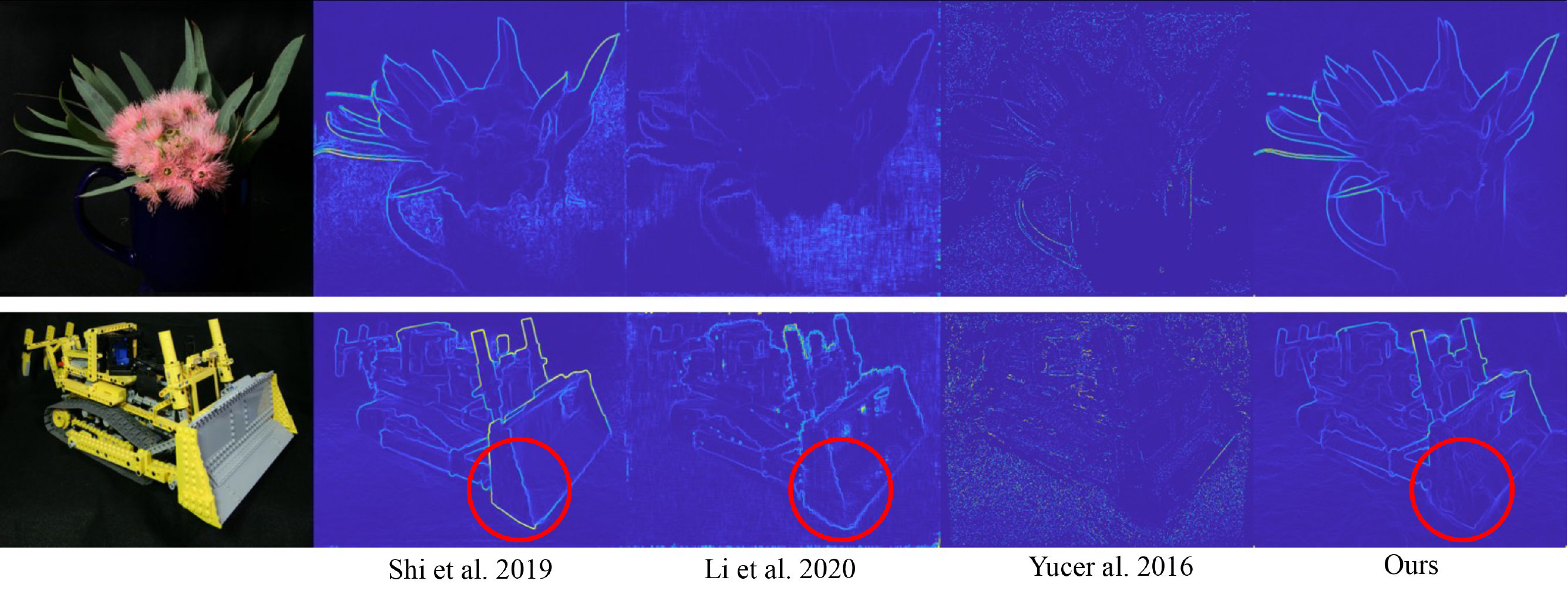}%
\vspace{-4mm}
\caption{Visualizing occlusion edges as gradients of disparity maps. \emph{Left to right:} Shi et al.~\cite{shi2019}, Li et al~\cite{li2020}, Yucer et al.~\cite{yucer2016}, and ours. \textit{Bottom row, red circle:}  the learning-based methods hallucinate a strong depth edge on the plow even though it is in contact with the black ground cloth at the same depth (supplemental Sec.~\ref{sec:appendix_additionaldiscussion}). Yucer et al.'s method fails in the absence of many views.} 
\label{fig:results-edges}
\end{figure*}

\begin{figure}[t]
\begin{minipage}[c]{0.45\linewidth}
\includegraphics[width=1.0\textwidth, trim={0.8cm 0 1.05cm 0},clip]{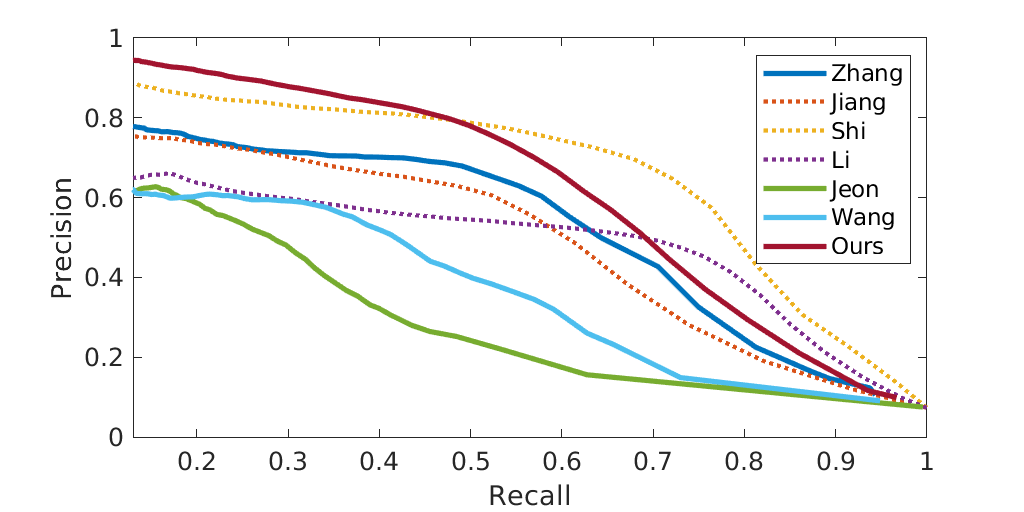}%
\end{minipage}
\begin{minipage}[c]{0.55\linewidth}
\caption{Average precision-recall curves of depth boundaries for all baseline algorithms (HCI dataset). Learning-based methods are shown as dotted lines. Our approach consistently outperforms traditional algorithms~\cite{jeon2015,wang16,zhang2016} and the learning-based method of Jiang et al.~\cite{jiang2018}, while outperforming Shi et al.~\cite{shi2019} and Li et al.~\cite{li2020} at medium-to-low recall rates. }
\label{fig:results-boundary-recall}
\end{minipage}
\vspace{-5mm}
\end{figure}

In Figure~\ref{fig:results-edges}, we visualize occlusion boundaries as depth gradients. While the learning-based methods of Shi et al.~and Li et al.~generating spurious boundaries in textureless regions, the approach of Yucer et al.~\cite{yucer2016} fails entirely in the absence of thousands of views. 
We also evaluate our edges quantitatively on four scenes from the synthetic HCI Light Field Dataset~\cite{honauer2016} via ground truth disparity maps for the central view (Fig.~\ref{fig:results-boundary-recall} and Tab.~\ref{table:results-synthetic}). Although our Q25 error is higher, our method has high boundary-recall precision, and a lower average mean-squared error than all baselines.

{
\newcolumntype{x}[1]{%
>{\centering\hspace{0pt}}p{#1}}%

\definecolor{best1}{rgb}{1, 0.8, 0} 
\definecolor{best2}{rgb}{0.78, 0.78, 0.78}
\definecolor{best3}{rgb}{0.79, 0.65, 0.44}

\newcommand{\gold}[1]{\colorbox{best1}{#1}}
\newcommand{\silver}[1]{\colorbox{best2}{#1}}
\newcommand{\bronze}[1]{\colorbox{best3}{#1}}

\setlength{\tabcolsep}{1.0pt}
\renewcommand{\arraystretch}{1.2}
\begin{table}[p]
\begin{center}
\caption{Quantitative comparison of our method and the baselines on the synthetic HCI light fields. The top three results are highlighted in \gold{gold}, \silver{silver} and \bronze{bronze}. MSE is the mean squared error; Q25 is the 25th percentile of the absolute error.}
\vspace{0.1mm}
\label{table:results-synthetic}
\resizebox{1.0\textwidth}{!}{%
\begin{tabular}{l|x{2.5em}|x{2.5em}|x{2.5em}|x{2.5em}|x{2.5em}|x{2.5em}||x{2.5em}|x{2.5em}|x{2.5em}|x{2.5em}|x{2.5em}|x{2.5em}||x{2.5em}|x{2.5em}|x{2.5em}|x{2.5em}|x{2.5em}|x{2.5em}|x{2.5em}|x{2.5em}|c}
\hline
\multirow{2}{4em}{Light Field} & \multicolumn{7}{c}{MSE $\times$ 100} & \multicolumn{7}{c}{Q25} & \multicolumn{7}{c}{Run time (s)} \\
\cline{2-22}
& \cite{jeon2015} & \cite{zhang2016} & \cite{jiang2018} & \cite{shi2019} & \cite{li2020} & \cite{wang16} & Ours & \cite{jeon2015} & \cite{zhang2016} & \cite{jiang2018} & \cite{shi2019} & \cite{li2020} & \cite{wang16} & Ours & \cite{jeon2015} & \cite{zhang2016} & \cite{jiang2018} & \cite{shi2019} & \cite{li2020} & \cite{wang16} & Ours \\
\hline
\textit{Sideboard} & 3.21 & \gold{1.02} & 1.96 & \bronze{1.12} & 1.89 & 13.3 & \silver{1.03} & \bronze{0.61} & 1.15 & \gold{0.37} & \silver{0.48} & 0.66 & 2.46 & 1.22 & 754 & 537 & 507 & \silver{72.3} & \bronze{77.1} & 635 & \gold{35.5} \\
\textit{Dino} & 1.73 & \gold{0.36} & 0.47 & \silver{0.43} & 3.28 & 4.19 & \bronze{0.45} & 1.07 & 1.40 & \gold{0.25} & \silver{0.31} & \bronze{0.50} & 2.02 & 0.85 & 805 & 531 & 500 & \silver{59.3} & \bronze{76.8} & 609 & \gold{37.7} \\
\textit{Cotton} & 12.5 & 1.81 & \bronze{0.97} & \silver{0.88} & 1.95 & 9.56 & \gold{0.70} & \bronze{0.50} & 1.01 & \gold{0.21} & \silver{0.36} & 0.59 & 2.30 & 0.74 & 748 & 530 & 500 & \bronze{79.8} & \silver{76.9} & 612 & \gold{34.0} \\
\textit{Boxes} & 16.0 & \bronze{7.90} & 11.6 & 8.48 & \gold{4.67} & 12.5 & \silver{7.52} & \bronze{0.75} & 1.64 & \gold{0.42} & \silver{0.69} & 0.78 & 2.21 & 1.41 & 736 & 541 & 491 & \silver{56.2} & \bronze{78.0} & 667 & \gold{34.3} \\
\hline
\textit{Average} & 8.37 & \bronze{2.77} & 3.75 & \silver{2.72} & 2.94 & 9.91 & \gold{2.43} & 0.73 & 1.3 & \gold{0.31} & \silver{0.46} & \bronze{0.63} & 2.25 & 1.05 & 761 & 535 & 500 & \silver{66.9} & \bronze{77.2} & 631 & \gold{35.4} \\
\hline
\end{tabular}
}
\end{center}
\end{table}
\setlength{\tabcolsep}{1.4pt}
}

\vspace{-0.2cm}
\paragraph{Diffusion Gradients as Self-supervised Loss.}
One way to think about bidirectional diffusion gradients is as a self-supervised loss function for depth edge localization. With this view, we compare its performance to \textit{multi-view reprojection error} --- a commonly used self-supervised loss in disparity optimization. 
We use the dense disparity maps $\hat{D}[\mathcal{P}_f]$ and $\hat{D}[\mathcal{P}_b]$ to warp all light field views onto the central view through an occlusion-aware inverse projection. 
A reprojection error map is calculated as the mean per-pixel L1 intensity error between the warped views and the central view. The offset direction at each point $p \in \mathcal{P}$ is then determined based on the disparity map that minimizes the reprojection error at the pixel location of $p$.  
Table~\ref{table:results-reproj-compare} evaluates the result of calculating $\mathcal{Q} = \{ p \pm \nabla I(p) ~\forall~p \in \mathcal{P}\}$ based on the reprojection error maps instead of our bidirectional diffusion gradients. Our method has consistently lower MSE, indicating better edge performance. This intuition is qualitatively confirmed by supplemental Figure \ref{fig:results-reproj-error}.

\begin{table}[t]
\begin{minipage}[c]{0.45\linewidth}
    \centering
    \resizebox{0.8\linewidth}{!}{
        \begin{tabular}{l c c c c}
        \toprule
        \multirow{2}{4em}{Light Field} & \multicolumn{2}{c}{MSE $\times$ 100} & \multicolumn{2}{c}{Q25} \\
        \cline{2-5}
        & Reproj & Ours & Reproj & Ours \\
        \midrule
        \textit{Sideboard} & 1.39 & \textbf{1.03} & \textbf{1.20} & 1.22 \\
        \textit{Dino} & 0.64 & \textbf{0.45} & \textbf{0.81} & 0.85 \\
        \textit{Cotton} & 1.04 & \textbf{0.70} & \textbf{0.68} & 0.74 \\
        \textit{Boxes} & 9.32 & \textbf{7.52} & 1.65 & \textbf{1.41} \\
        \midrule
        \textit{Average} & 3.10 & \textbf{2.43} & 1.08 & \textbf{1.05}\\
        \bottomrule
        \end{tabular}
    }
\end{minipage}
\begin{minipage}[c]{0.55\linewidth}
    \caption{Evaluating disparity maps with depth edges identified via reprojection error and via our approach of diffusion gradients on the synthetic HCI dataset. MSE is the mean squared error; Q25 is the 25th percentile of absolute error.}
    \label{table:results-reproj-compare}
\end{minipage}
\vspace{-0.25cm}
\end{table}

\section{Discussion}
\label{sec:discussion}

\begin{figure}[b]
\centering
\vspace{-0.5cm}
\includegraphics[width=0.8\textwidth]{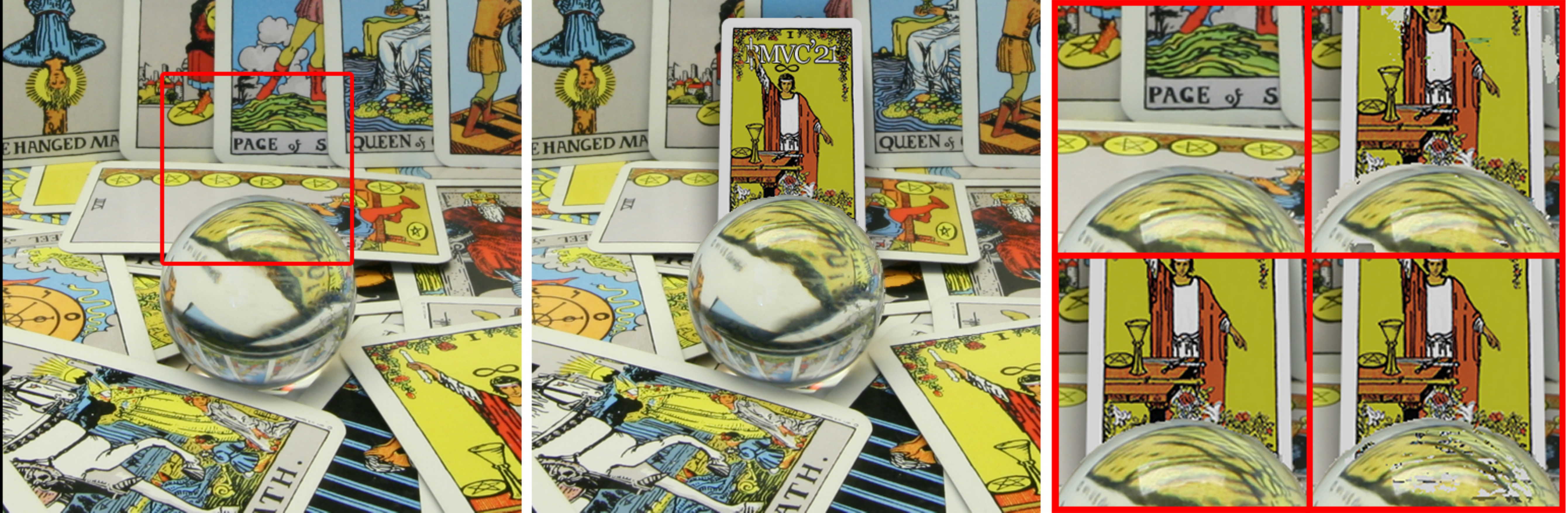}%
\vspace{-0.25cm}
\caption{Adding a BMVC'21 tarot card to the scene. \emph{Left:} input  scene. \emph{Center:} Our editing results. \emph{Right, clockwise from top-left}: Detail of the unmodified light field image, Zhang et al.~\cite{zhang2016}'s editing result, Shi et al.~\cite{shi2019}'s editing result, and our result with fewer artifacts. }
\label{fig:editing-tarot}
\vspace{-0.25cm}
\end{figure}

\paragraph{Light Field Editing.} As our method generates accurate depth edges that allow visibility to be handled correctly, our depth allows simple object insertion with few artifacts (Figs.~\ref{fig:editing1} and~\ref{fig:editing-tarot}).

\vspace{-0.2cm}
\paragraph{Errors.} Our method has consistently lower mean squared error (MSE), but suffers a higher number of erroneous pixels (Q25). As Q25 measures the first quantile of absolute error, 
this indicates that baseline methods must have more outliers: the errors that they do have must be considerably large. 
This intuition is confirmed by visualizing the absolute error (supplemental Fig.~\ref{fig:results-error}) which shows regions of large error around occlusion boundaries for the baseline methods.

We also observe evidence supporting our initial comment that supervised deep-learning-based methods can overfit. When tested on real-world light fields, methods trained only on the HCI dataset~\cite{li2020} produce artifacts along depth edges (Fig.~\ref{fig:results-real}). More varied training data or more sophisticated data augmentation methods could be employed here. Our method is not susceptible to these problems, producing comparable output on both synthetic and real-world light fields.

\vspace{-0.2cm}
\paragraph{Conclusion.}
\label{sec:conclusion}
Estimating occlusion-accurate depth maps from light fields is useful for scene editing and AR applications. Our approach is based around a bidirectional diffusion process that can disambiguate depth from color edges and estimate a correct depth edge offset to provide accurate gradient information for diffusion. We also contribute a faster method to find sub-pixel disparity labels at a sparse set of points via an entropy-based depth refinement process. The effectiveness of this strategy is shown with results on synthetic and real world light fields, producing competitive or better mean squared error accuracy while being significantly faster than other non-learning-based methods.


\section*{Acknowledgments}
Numair acknowledges an Andy van Dam PhD Fellowship, and James acknowledges a gift from Cognex. Min H. Kim acknowledges the support of Korea NRF grant (2019R1A2C3007229)

\bibliography{bibliography}

\clearpage
\appendix

\section{Supplemental Material}

\begin{figure}[t]
\centering
\includegraphics[width=\textwidth]{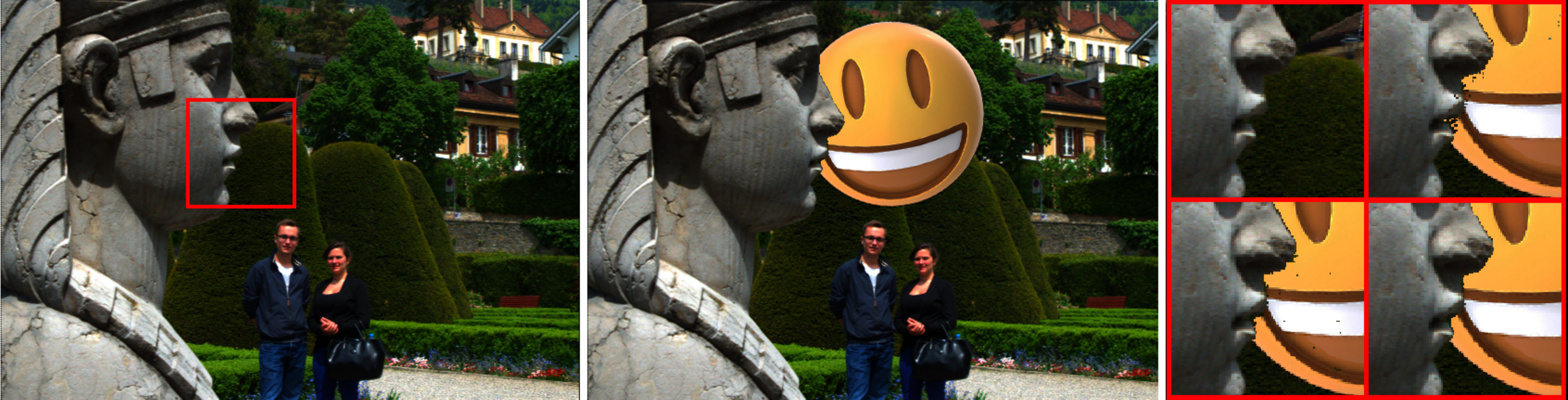}%
\vspace{-0.25cm}
\caption{Additional light field editing result. \emph{Left:} input scene. \emph{Center:} Our editing results. \emph{Right, clockwise from top-left}: Detail of the unmodified light field image, Zhang et al.~\cite{zhang2016}'s editing result, Shi et al.~\cite{shi2019}'s editing result, and our result with fewer artifacts.}
\label{fig:editing-sphynx}
\end{figure}

We include additional discussion covering the benefits over naive diffusion, consistency over views within the 4D light field, tolerance to depth label errors and edge blur, details of dataset preprocessing, and an example of textures within dark backgrounds in the Stanford dataset (Section \ref{sec:appendix_additionaldiscussion}). Next, we present error maps comparing reprojection loss versus our bidirectional diffusion approach (Section \ref{sec:appendix_selfsupervisedloss}), and error maps versus ground truth for the HCI dataset (Section \ref{sec:appendix_errormaps}). Finally, we show additional qualitative results on the Stanford dataset (Section \ref{sec:appendix_expandedresults}) and an additional editing example (Figure \ref{fig:editing-sphynx}).

\subsection{Additional Discussion}
\label{sec:appendix_additionaldiscussion}

\begin{figure}[h]
\centering
\includegraphics[clip,trim={0cm 19cm 0cm 0cm},width=0.4\linewidth]{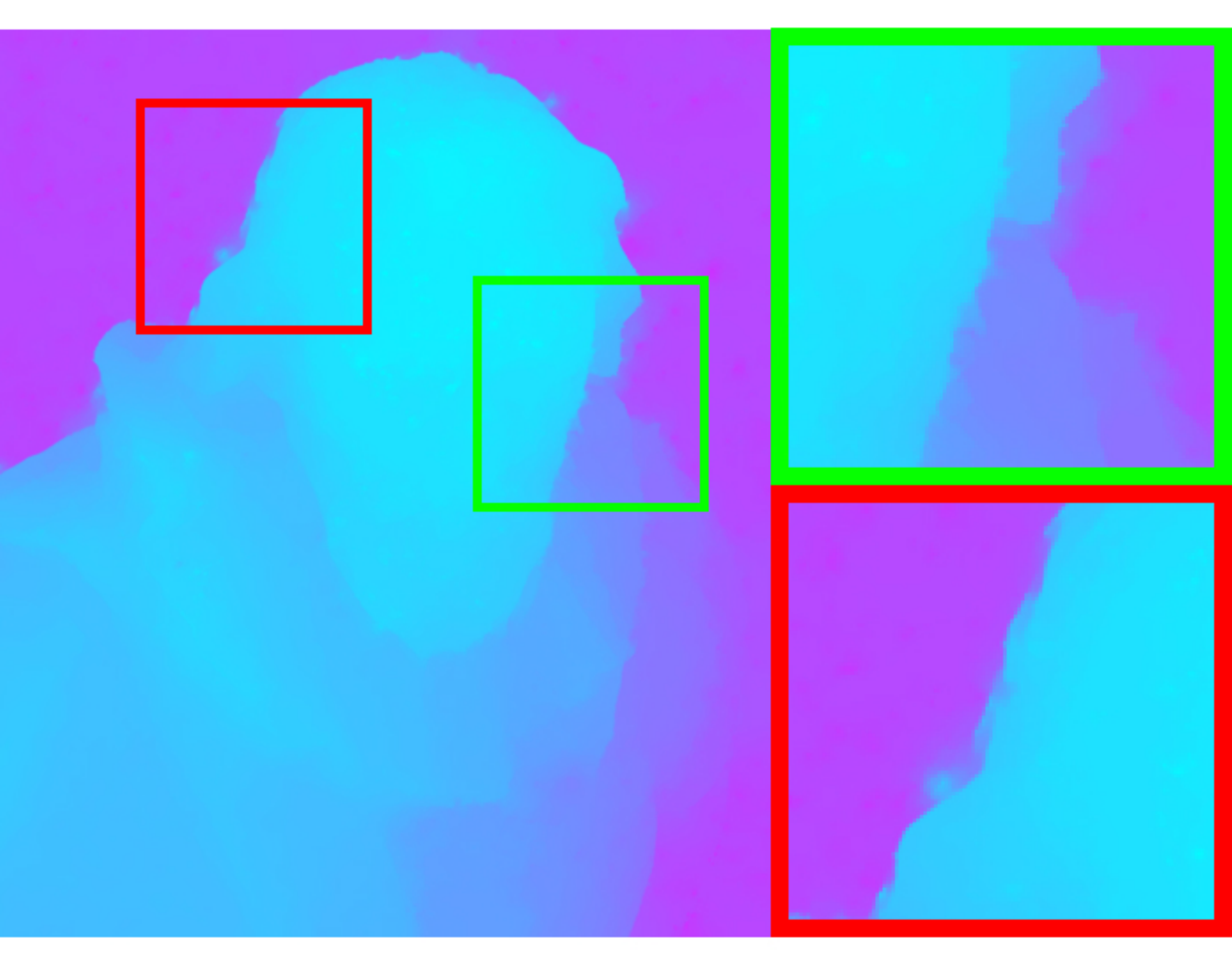}
\includegraphics[clip,trim={0cm 0cm 0cm 19.5cm},width=0.4\linewidth]{Figures/naive.pdf}
\caption{\emph{Left}: Na{\"i}vely diffusing disparity labels causes artifacts around edges due to ambiguity in the localization of labels around edges. \emph{Right}: Estimating the diffusion gradient removes this ambiguity and yields sharp depth edges.}
\label{fig:epi-naive-diffusion}
\end{figure}

\paragraph{Naive Diffusion.} 
In Figure~\ref{fig:epi-naive-diffusion}, we demonstrate visually that naively diffusing disparity labels can be problematic because edge localization is ambiguous.

\begin{figure}[h]
\centering
\includegraphics[width=1.0\linewidth]{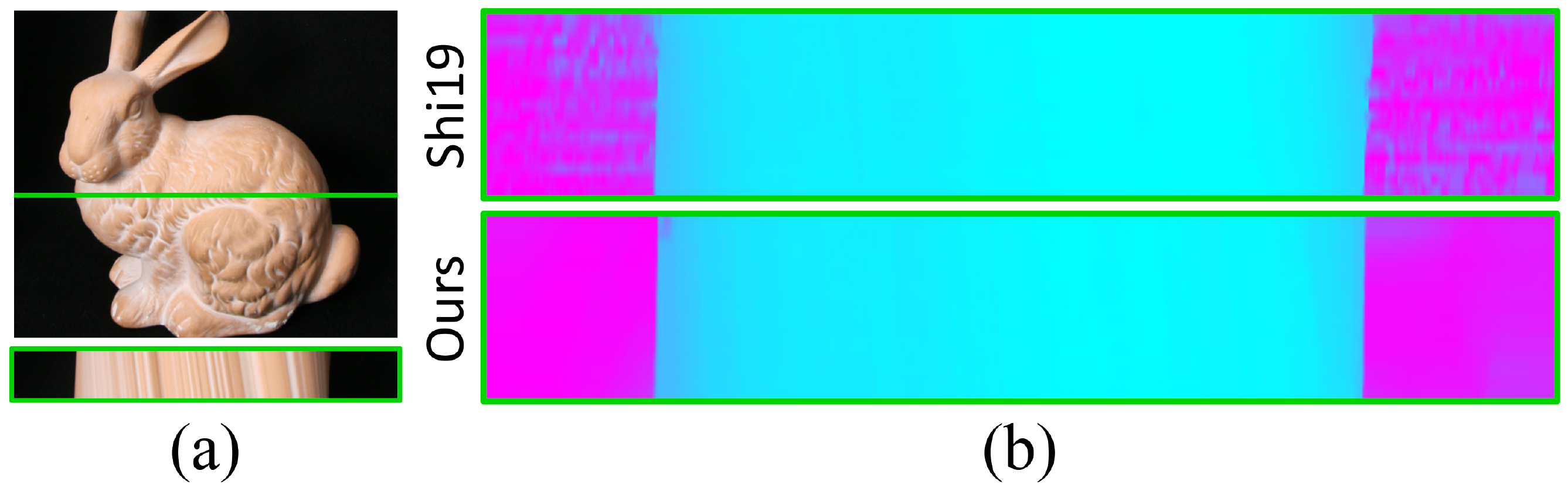}
\caption{\textbf{(a)} We visualize depth consistency for the highlighted epipolar line. \textbf{(b)} Our results are more consistent  than Shi et al.~\cite{shi2019} across views (EPIs are scaled vertically for clarity).}
\label{fig:results-multiview}
\end{figure}

\paragraph{Multi-view Depth and Error.}
As ground truth disparity is only provided for the central view of the HCI data set, and as the Stanford data set has no ground truth depth, we did not include quantitative error evaluation across `4D' views. Qualitatively, our method tends to produce results that are consistent across views (Fig.~\ref{fig:results-multiview}).

\begin{figure}[h]
\centering
\vspace{-3mm}
\includegraphics[width=0.8\linewidth]{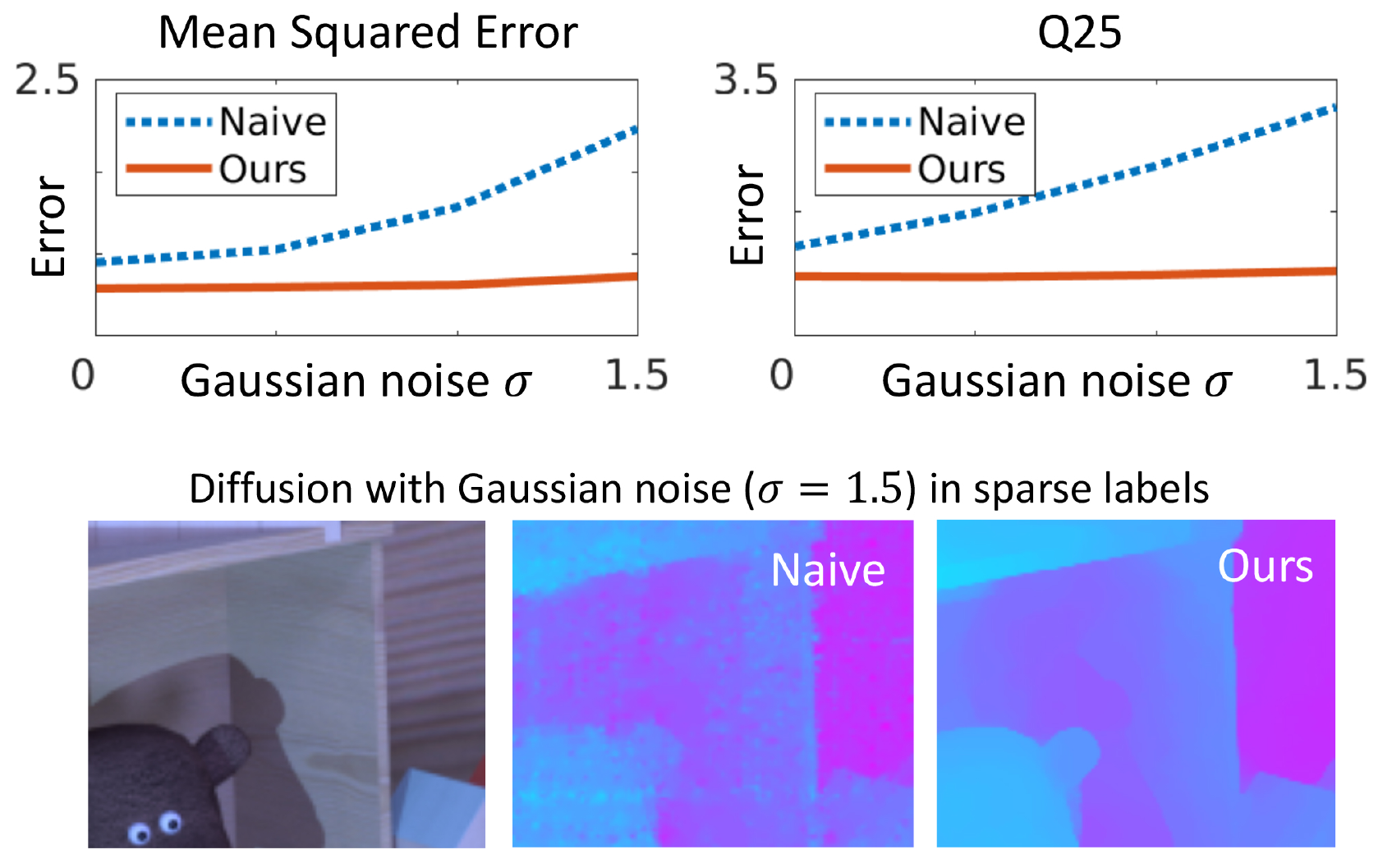}
\vspace{-3mm}
\caption{Robustness of our method to noise in disparity labels (\textit{Dino} light field; we compare with naive diffusion.).}
\label{fig:noisy-results}
\end{figure}

\begin{figure}[h]
\centering
\vspace{-3mm}
\includegraphics[width=0.7\linewidth]{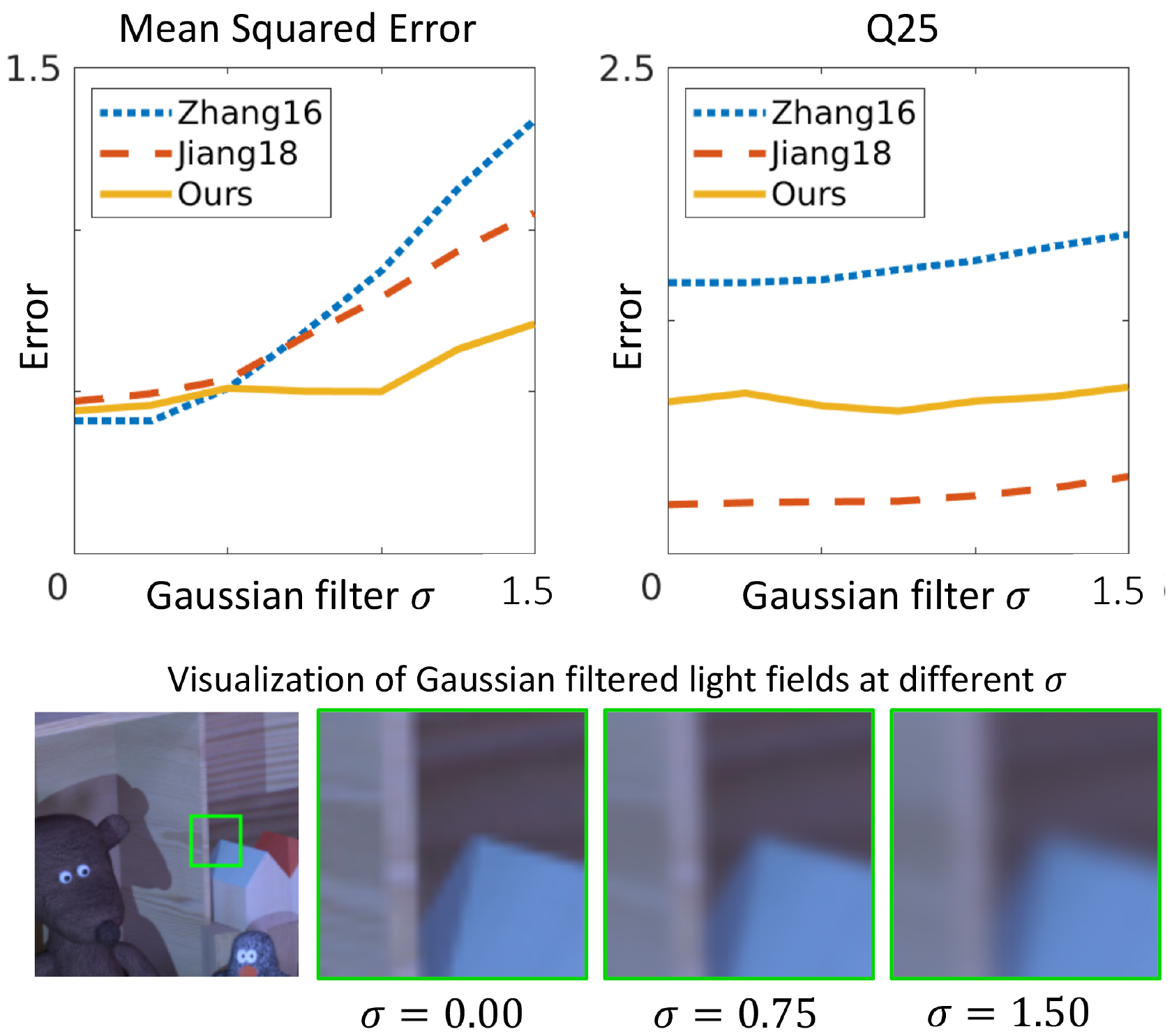}
\vspace{-3mm}
\caption{Robustness of our method to low-gradient edges (\textit{Dino} light field; we compare to the methods of Zhang et al.~\cite{zhang2016} and Jiang et al.~\cite{jiang2018} which have the best MSE and Q25 performance on this light field, respectively).}
\label{fig:filtered-results}
\end{figure}

\paragraph{Disparity Noise and Blur Tolerance.} 
To show our robustness, we evaluate our method on noisy disparity labels (Fig.~\ref{fig:noisy-results}) and low-gradient edges (Fig.~\ref{fig:filtered-results}). Our method provides greater robustness to disparity errors than naive diffusion, and provides greater robustness via MSE to low-gradient (or blurry) edges than two learning-based baselines.

\paragraph{Lenslet Distortion and EPFL Lytro Dataset.} The Lytro light fields in the EPFL dataset are provided decoded as MATLAB files. In general, while our method can handle small amounts of distortion, the EPI-based edge detection stage expectedly fails when EPI features are no longer linear. This is true for the edge views of Lytro light fields. As such, we only use the central 7$\times$7 views of the EPFL scenes for all experiments.

\paragraph{Black Backgrounds and Stanford Dataset.} Our EPI edge detector aggregates information from all three channels in CIE LAB color space, which allows it to detect even faint edges. Thus, it captures the subtle background texture on the black cloth in the Stanford dataset examples of single objects; typically, this detail is not visible to the naked eye. This feature of our work also explains why we do not incorrectly detect false edges in the Lego Technic Plow scene, as shown in Figure \ref{fig:results-edges} of the main paper.
\begin{figure}[h]
\begin{center}
\includegraphics[width=1.0\linewidth]{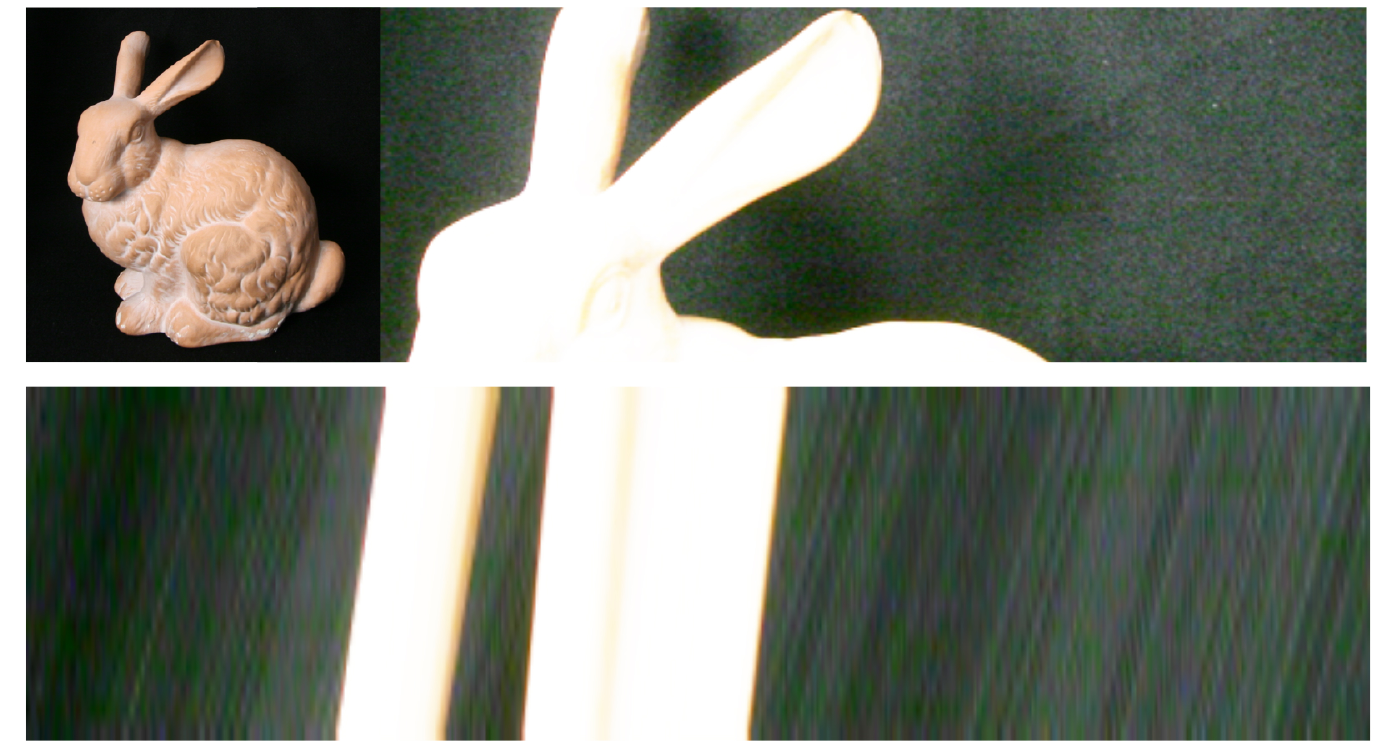}
\end{center}
\vspace{-0.5cm}
\caption{\textit{Top:} On the Stanford Bunny scene, enhanced image contrast shows the texture of the cloth in the seemingly black background. \textit{Bottom:} In EPI space (scaled vertically for clarity) the texture appears as sloped lines, providing background disparity to methods that can exploit this subtle information.}
\label{fig:sparse-texture}
\end{figure}

\subsection{Expanded Results}
\label{sec:appendix_expandedresults}

\begin{figure*}[tp]
\centering
\includegraphics[width=\textwidth]{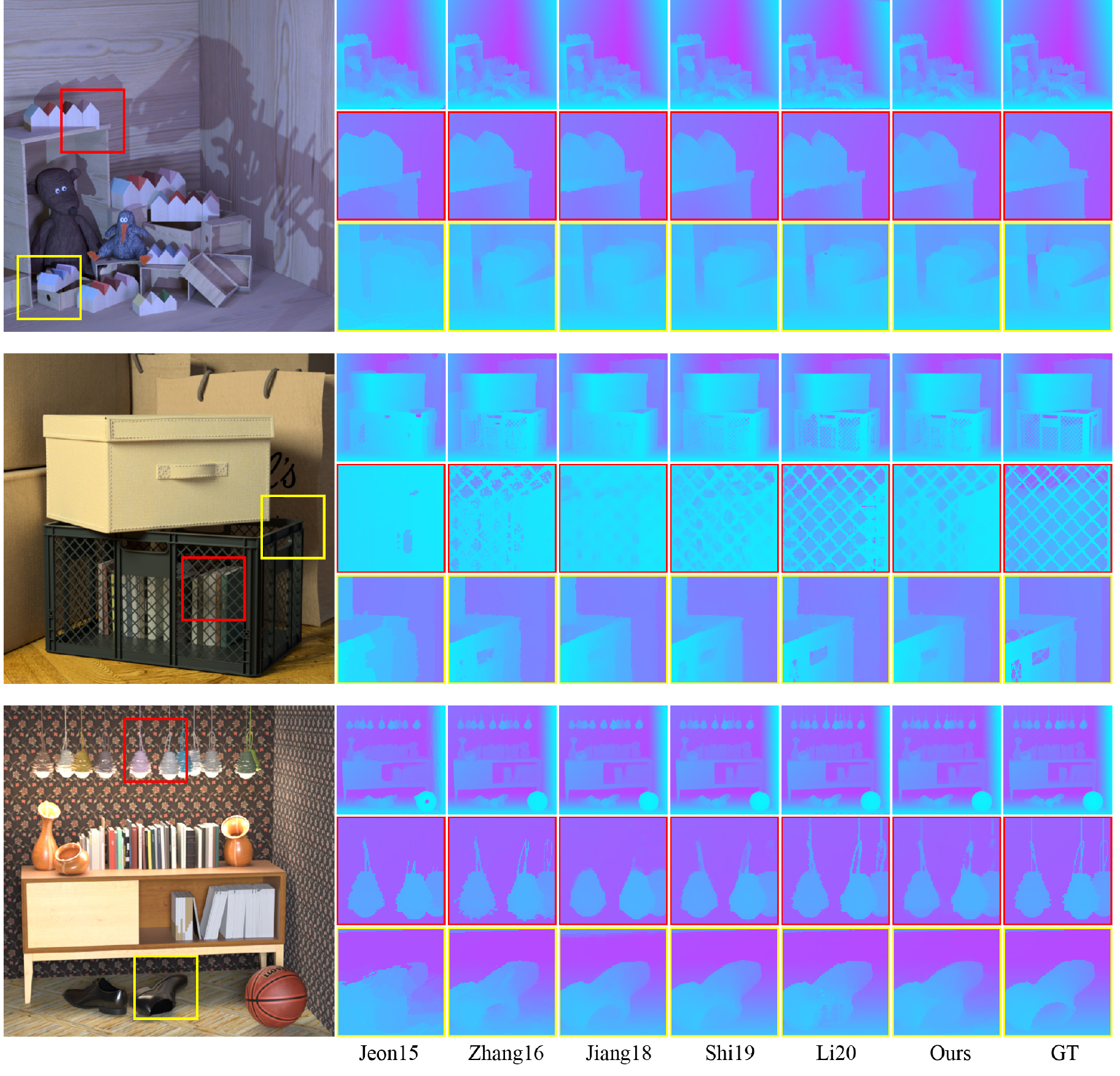}
\vspace{-0.25cm}
\caption{Results on the synthetic light fields of the HCI dataset. \emph{Left to right:} Jeon et al.~\cite{jeon2015}, Zhang et al~\cite{zhang2016}, Jiang et al.~\cite{jiang2018}, Shi et al.~\cite{shi2019}, Li et al.~\cite{li2020}, our method, and finally, the ground truth. Qualitatively, our results are comparable to the learning-based baselines~\cite{li2020, shi2019, jiang2018} with fewer extreme errors around edges.} 
\label{fig:results-hci}
\end{figure*}

\begin{figure*}[t!]
\centering
\includegraphics[width=\textwidth]{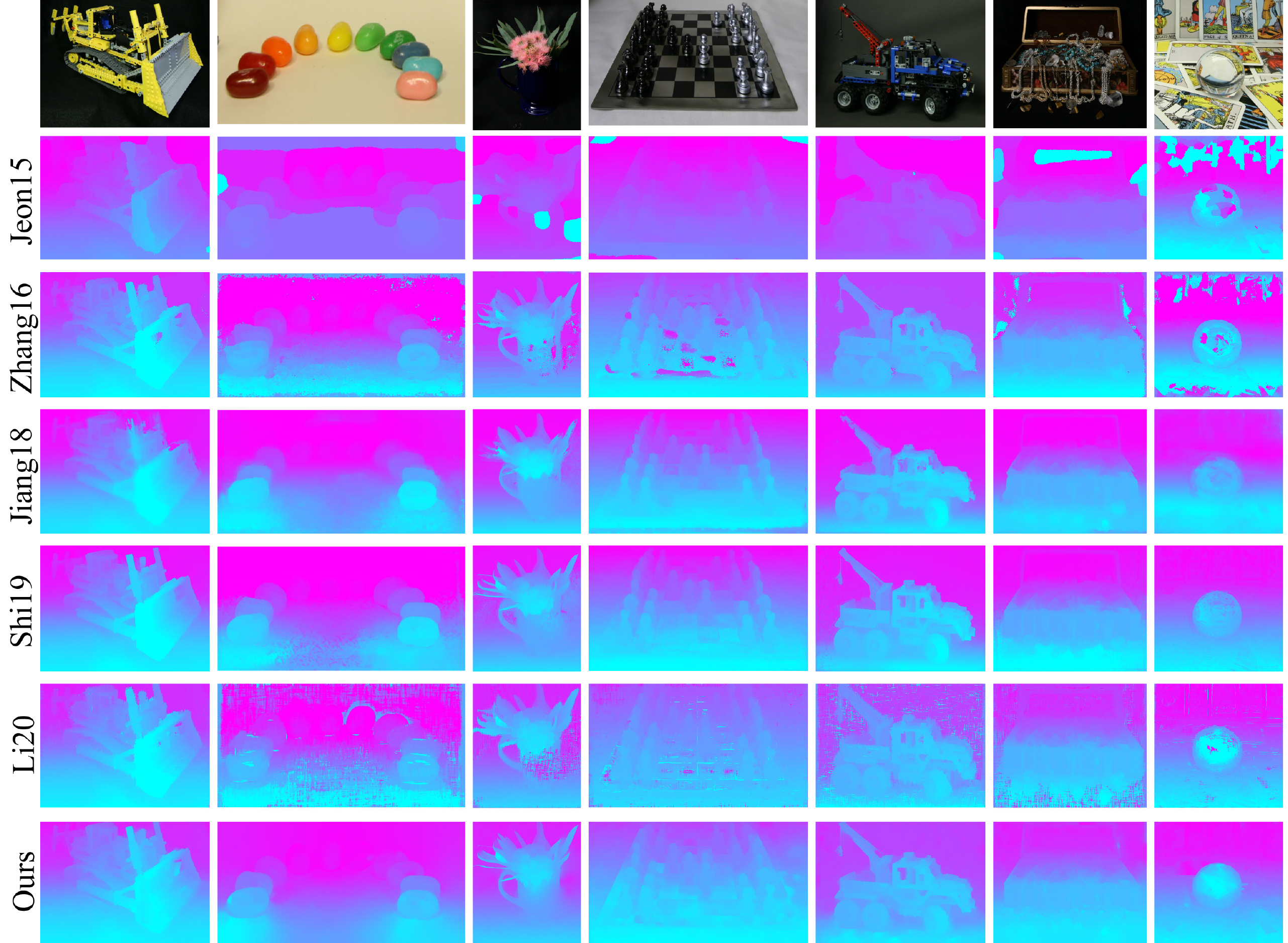}
\caption{Results on light fields from the Stanford dataset. \emph{Top to bottom:} Jeon et al.~\cite{jeon2015}, Zhang et al~\cite{zhang2016}, Jiang et al.~\cite{jiang2018}, Shi et al.~\cite{shi2019}, Li et al.~\cite{li2020} and our method. } 
\label{fig:results-expanded-stanford}
\end{figure*}

We present qualitative results on the HCI dataset in Figure~\ref{fig:results-hci}, and expanded results on the real-world light fields of the Stanford dataset in Figure~\ref{fig:results-expanded-stanford}. Our method produce stronger depth edges compared to the baselines, and our smoothness regularization (Equation \ref{eqn:diffusion-data-final}, main paper) leads to fewer artifacts in textureless regions.

\subsection{Diffusion Gradients as Self-supervised Loss}
\label{sec:appendix_selfsupervisedloss}

\begin{figure}[t]
    \centering
    \includegraphics[width=0.95\linewidth]{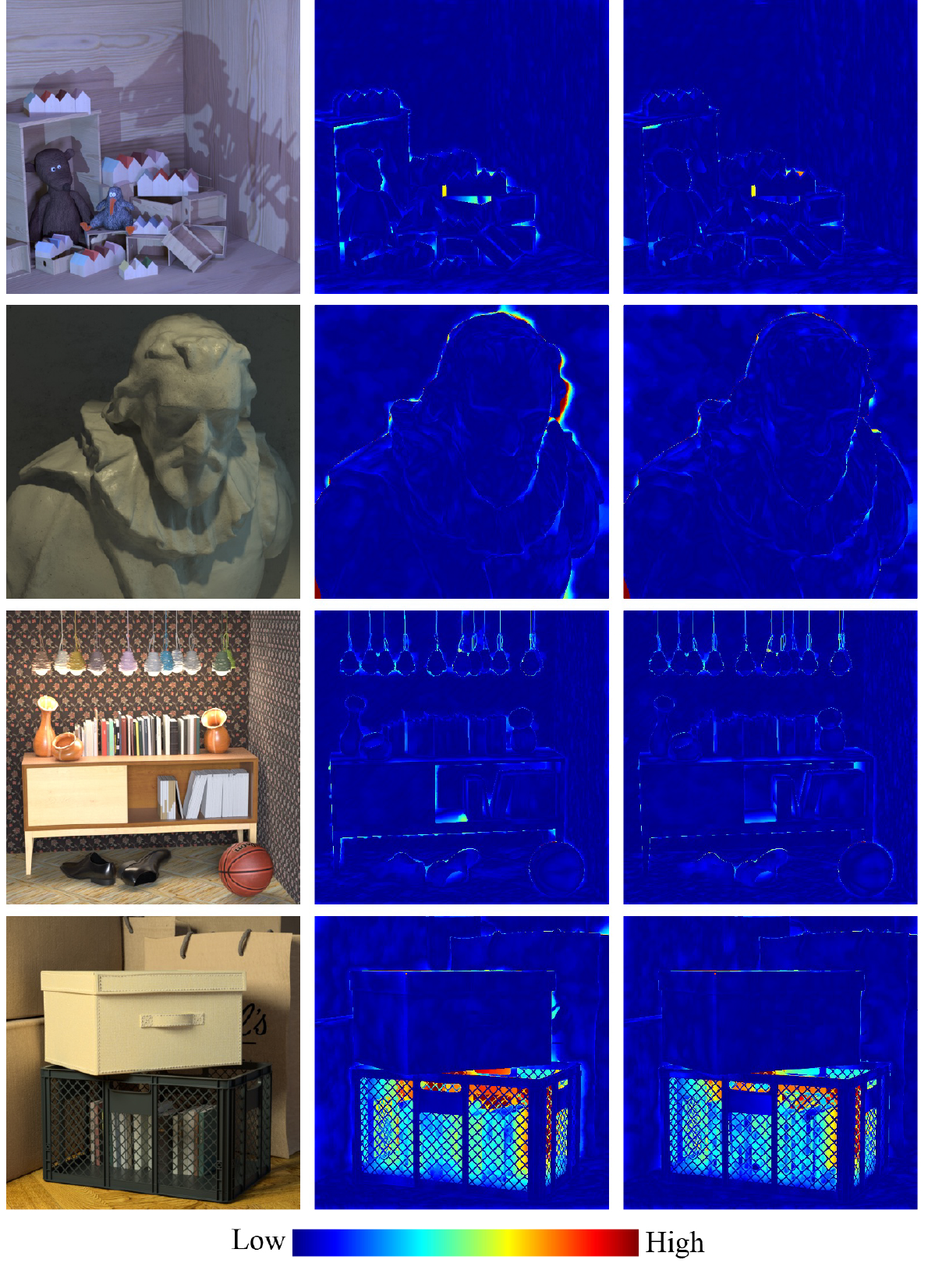}
    \caption{Multiview reprojection error (\emph{center}) as self-supervised loss for depth edge localization, compared to our bidirectional diffusion gradients (\emph{right}). We show absolute disparity error. Our method has lower error around edges.} 
    \label{fig:results-reproj-error}
\end{figure}

As in main paper Section \ref{sec:experiments}, we compare our method to a reprojection error loss. In Figure~\ref{fig:results-reproj-error}, to complement the quantitative MSE numbers in the main paper, we demonstrate the qualitative improvement from our bidirectional diffusion gradient approach in comparison.

\subsection{Error Maps}
\label{sec:appendix_errormaps}

\begin{figure*}[t!]
\centering
\includegraphics[width=0.8\textwidth]{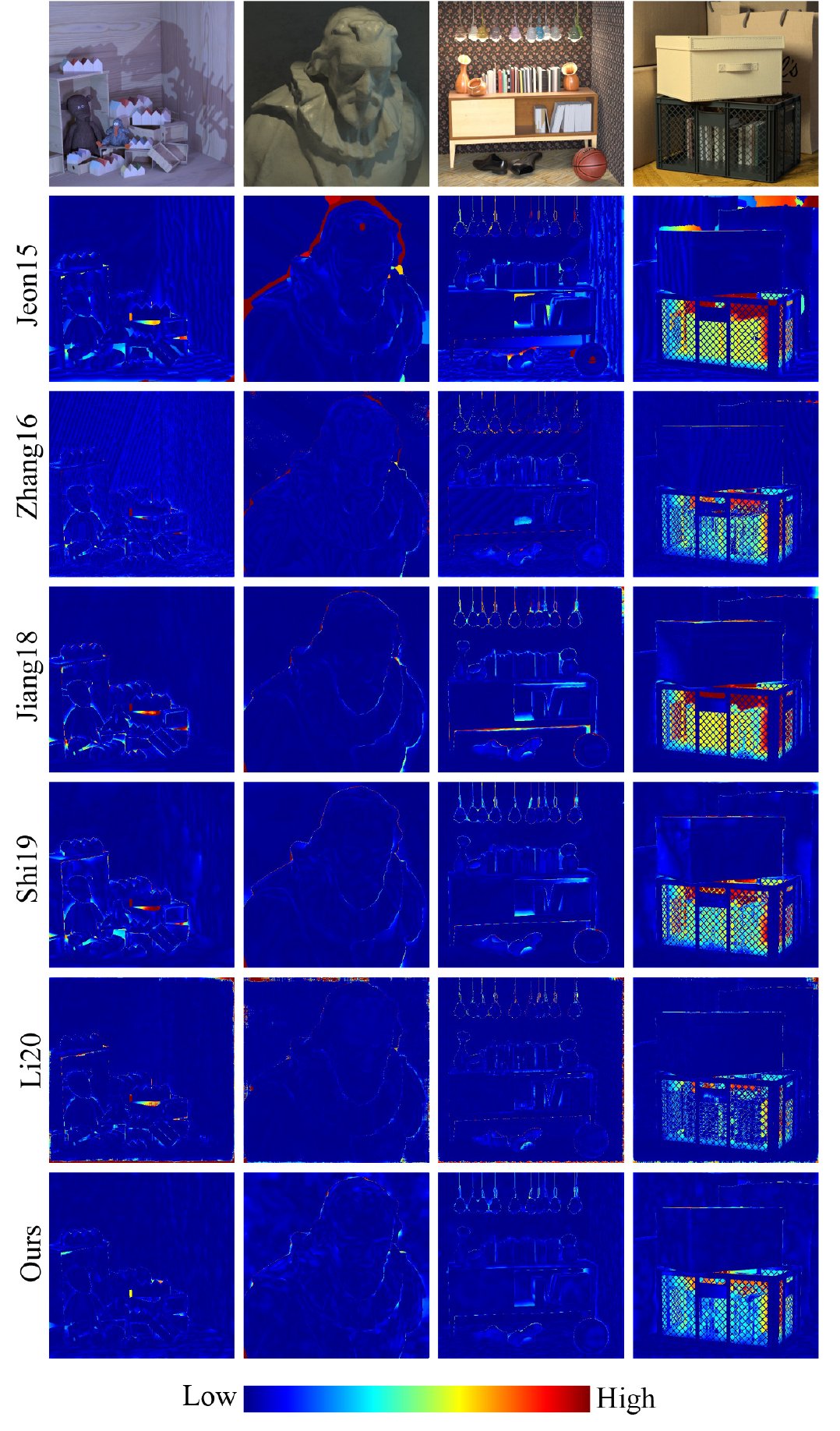}
\caption{A visualization of the absolute disparity error for all baselines. \emph{Top to bottom:} Jeon et al.~\cite{jeon2015}, Zhang et al~\cite{zhang2016}, Jiang et al.~\cite{jiang2018}, Shi et al.~\cite{shi2019}, Li et al.~\cite{li2020}, and our method.} 
\label{fig:results-error}
\end{figure*}

We visualize the absolute disparity error of all baselines and our method in Figure~\ref{fig:results-error}. The baseline methods produce larger errors around depth edges compared to our approach. This can be seen in the fewer regions of red for our method compared to the baselines. The corresponding dense disparity maps are shown in Figure~\ref{fig:results-hci}. Qualitatively, our results are comparable to the learning-based baselines~\cite{li2020, shi2019, jiang2018} with fewer extreme errors around edges. 

\end{document}